\def\year{2022}\relax
\documentclass[letterpaper]{article} %
\usepackage{aaai22}  %
\usepackage{times}  %
\usepackage{helvet}  %
\usepackage{courier}  %
\usepackage[hyphens]{url}  %
\usepackage{graphicx} %
\usepackage{natbib}  %
\usepackage{caption} %
\DeclareCaptionStyle{ruled}{labelfont=normalfont,labelsep=colon,strut=off} %
\usepackage{algorithm}
\usepackage{algorithmic}

\usepackage{newfloat}
\usepackage{listings}
\floatstyle{ruled}
\newfloat{listing}{tb}{lst}{}
\floatname{listing}{Listing}

\usepackage{ulem}
\usepackage{amssymb}
\usepackage[switch]{lineno}
\usepackage{multirow}
\usepackage{amsmath}
\usepackage{caption}
\usepackage{array}
\usepackage{booktabs}
\usepackage{footnote}
\usepackage{graphicx, subfig}
\usepackage{color}
\usepackage{subfig}
\usepackage{bbding}
\usepackage{harpoon}

\title{UniMS: A Unified Framework for Multimodal Summarization with Knowledge Distillation}
\author{
    Zhengkun Zhang\textsuperscript{\rm 1}\footnote{Work is done at the internship of Noah's Ark Lab, Huawei Technologies.},
    Xiaojun Meng\textsuperscript{\rm 2},
    Yasheng Wang\textsuperscript{\rm 2},
    Xin Jiang\textsuperscript{\rm 2},
    Qun Liu\textsuperscript{\rm 2},
    Zhenglu Yang\textsuperscript{\rm 1}\footnote{Corresponding authors.}
}
\affiliations{
    \textsuperscript{\rm 1}TKLNDST, CS, Nankai University, China,
    \textsuperscript{\rm 2} Noah's Ark Lab, Huawei Technologies\\
    
    zhangzk2017@mail.nankai.edu.cn, \\
    \{xiaojun.meng, wangyasheng, Jiang.Xin, qun.liu\}@huawei.com, \\
    yangzl@nankai.edu.cn
}

\usepackage{bibentry}

\begin{document}
\normalem

\maketitle

\begin{abstract}
With the rapid increase of multimedia data, a large body of literature has emerged to work on multimodal summarization, the majority of which target at refining salient information from textual and visual modalities to output a pictorial summary with the most relevant images. Existing methods mostly focus on either extractive or abstractive summarization and rely on qualified image captions to build image references. We are the first to propose a \textbf{U}nified framework for \textbf{M}ultimodal \textbf{S}ummarization grounding on BART, \textbf{UniMS}, that integrates extractive and abstractive objectives, as well as selecting the image output. Specially, we adopt knowledge distillation from a vision-language pretrained model to improve image selection, which avoids any requirement on the existence and quality of image captions. Besides, we introduce a visual guided decoder to better integrate textual and visual modalities in guiding abstractive text generation. Results show that our best model achieves a new state-of-the-art result on a large-scale benchmark dataset. The newly involved extractive objective as well as the knowledge distillation technique are proven to bring a noticeable improvement to the multimodal summarization task.
\end{abstract}

\section{Introduction}
\begin{figure}[t]
\centering
\includegraphics[width=.5\textwidth]{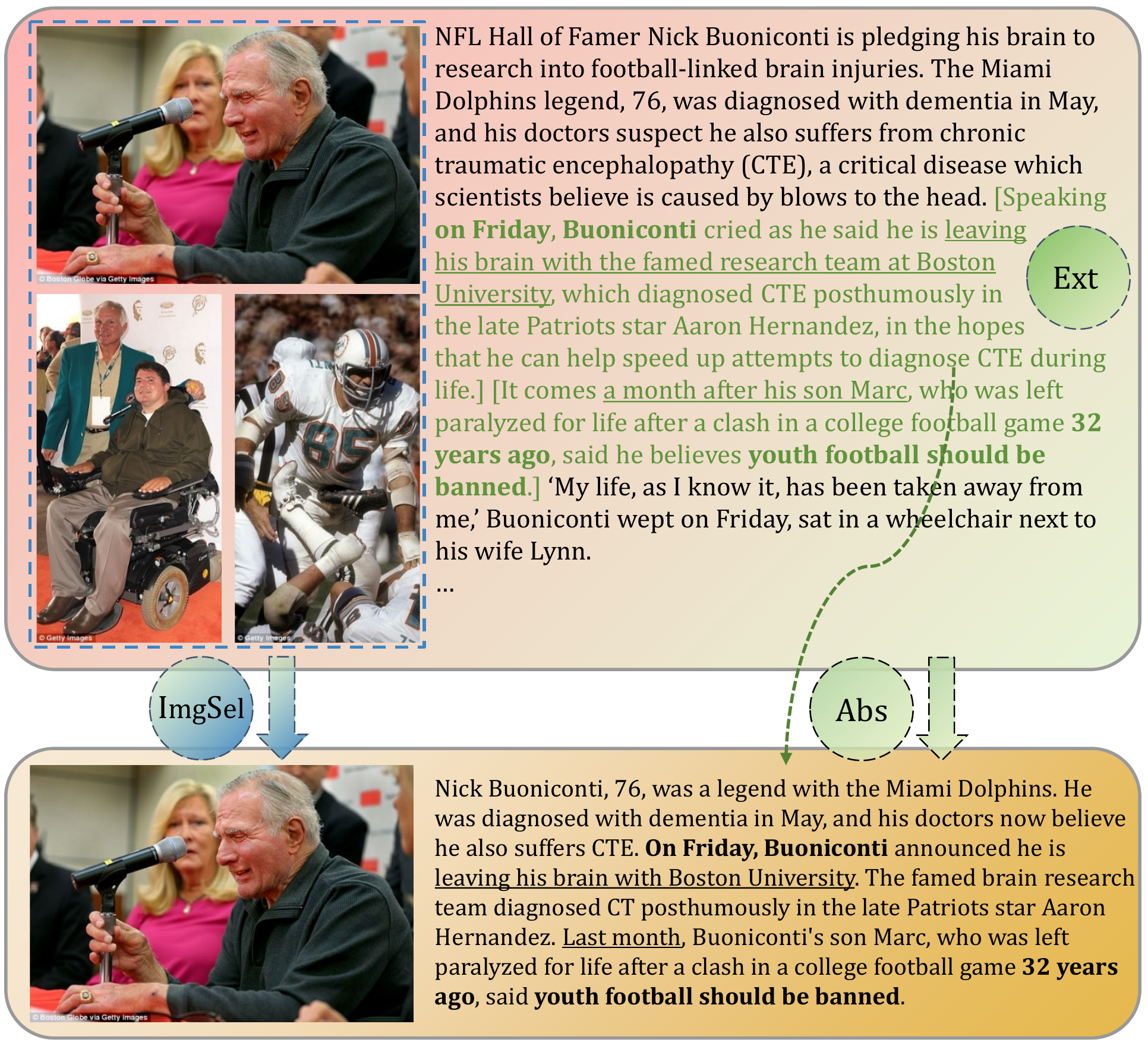}
\caption{
The illustration of multimodal summarization task. Our unified framework is able to achieve extractive and abstractive summarization objectives, as well as the image selection objective with no need of using image captions.
}
\label{fig:intro}
\end{figure}

Existing researchers \cite{DBLP:journals/air/Srihari94,DBLP:journals/spm/HeD17,DBLP:journals/pami/BaltrusaitisAM19} have evidenced that in order to understand the world around us, it needs to be able to interpret and reason from multimodal information.
Recently proposed Multimodal Summarization with Multimodal Output \cite{DBLP:conf/emnlp/ZhuLL0ZZ18} (MSMO) that condenses long multimodal news to a short pictorial version, as shown in Fig.~\ref{fig:intro}, is a new fashion of summarization. This method improves users' satisfaction with the ability to quickly grasp news highlights, and brings both research and commercial value in the face of information explosion.

A large amount of methods \cite{DBLP:conf/emnlp/ZhuLL0ZZ18, DBLP:conf/aaai/ZhuZZLZL20, DBLP:conf/aaai/ZhangWSY21} are proposed for improving multimodal summarization. They serve as baselines for the large-scale benchmark dataset collected by \citet{DBLP:conf/emnlp/ZhuLL0ZZ18}, and provide insights for recent work on multimodal summarization. In order to obtain the pictorial summary including a piece of condensed text with images as shown in Fig.~\ref{fig:intro}, existing methods typically consist of two objectives.

Firstly, as a critical modality of the pictorial summary, the objective of abstractive text summarization plays an important role in improving multimodal summarization. Recent advances in denoising autoencoder for pretraining sequence-to-sequence models such as BART \cite{DBLP:conf/acl/LewisLGGMLSZ20} have been shown to capture many language relevant facets for downstream tasks, which leads to a large improvement on text summarization metrics. We believe the existing high-capacity language model benefits summarization remarkably even in a multimodal setting. However, the original BART model has no support for multimodal input and output. Therefore, we are motivated to integrate textual and visual modalities into the BART model and extend its architecture to further improve the multimodal summarization task.

Next, to leverage the visual modality and improve the quality of model-selected image, \citet{DBLP:conf/aaai/ZhuZZLZL20} proposes to incorporate image references into the training process, and jointly consider summary generation and image selection as training targets. Due to the lack of image references, \citet{DBLP:conf/aaai/ZhuZZLZL20} proposes a ROUGE-ranking method using the similarity of image caption and text reference to rank images, which hence builds the pseudo image reference for training. It assumes the image caption has already contained condensed textual information of visual modalities, which semantically matches with the corresponding image. However, this assumption largely depends on the quality of the image caption instead of the image content itself. To sum up, the effectiveness of this ROUGE-ranking method relies on the presence of high-quality image captions. Unfortunately, the image captions are not often qualified or even not present in a multitude of irregular multimedia data.

Thus, we propose to distill knowledge from a pretrained vision-language teacher model (\emph{e.g.,} CLIP) to help our model on learning the relevance rank of images in the given textual context without any requirement on the existence and quality of image captions. Specially, given the recent process in contrastive vision-language pre-training \cite{DBLP:conf/icml/RadfordKHRGASAM21}, we are motivated to use the cosine similarity of text references and images to represent text-image content relevance, and use this relevance to build pseudo image references. As far as we know, distilling knowledge from vision-language models to mentor multimodal summarization remains unexplored. 

Considering the above mentioned abstractive and image selection objectives, we design \textbf{UniMS}: a \textbf{Uni}fied framework for \textbf{M}ultimodal \textbf{S}ummarization grounding on BART, which integrates inputs of both textual and visual modalities along with multiple multimodal objective functions in a unified multitask framework. We modify the BART architecture separately on its encoder and decoder. In the encoder, we distill the knowledge from a pretrained vision-language teacher model to guide our encoder on selecting images. We additionally introduce a new extractive objective, together with the image selection objective, as multimodal supervised signals for our encoder. It could potentially reduce the modality-bias problem~\cite{DBLP:conf/aaai/ZhuZZLZL20} and improve the understanding of multi-modalities for our encoder, which eventually benefits the multitask abilities. In the decoder, we adopt visual guided attention to better integrate both modalities to achieve the abstractive objective.

Overall, our unified multitask framework extends the BART model by enabling both extractive and abstractive summarization, as well as image selection as output. To our knowledge, \textbf{UniMS} is the first unified framework including three objectives of multimodal summarization. Experimental results show that our framework achieves new state-of-the-art results in all tasks. Our contributions in this paper are three-fold:

1. We propose a unified multimodal summarization framework with an encoder-decoder multitask architecture on top of BART, which simultaneously outputs extractive and abstractive summaries, and image selection results.

2. Our framework adopts knowledge distillation to improve image selection without any requirement on the existence and quality of image captions. We further introduce the extractive objective in the encoder and visual guided attention in the decoder to better integrate both textual and visual modalities in the conditional text generation.

3. Our unified method achieves a new state-of-the-art result of multimodal summarization in all the subtasks, i.e., extractive, abstrastive, and image selection.

\begin{figure*}
\centering
\includegraphics[width=.9\textwidth]{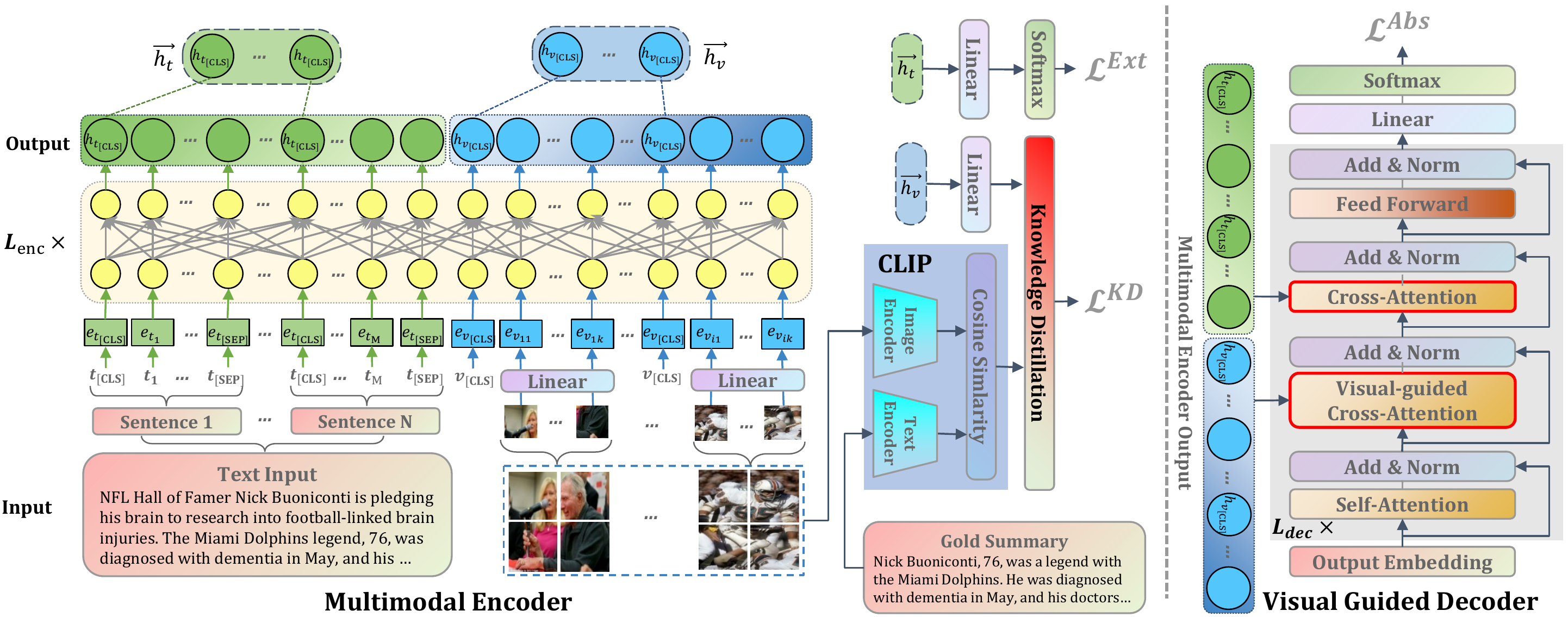}
\caption{
Overview of our proposed UniMS framework.
At first, we extend BART encoder to multimodal encoder by embedding visual features with linear projection as additional input.
To enhance the multimodal encoder's ability to understand multimodal input, we further introduce extractive text reference and image reference via knowledge distillation to guide our encoder. Besides, the proposed visual guided decoder separately attends to the visual and the textual output of the multimodal encoder.
}
\label{fig:model}
\end{figure*}

\section{Related Work}

Multimodal summarization takes two or more modalities of data as input, and outputs either single-modal or multimodal summary. The input modality could contain documents, images, audios and videos from the rich multimedia information \cite{DBLP:conf/emnlp/LiZMZZ17}, which enables it to make use of more information than traditional text summarization. \citet{DBLP:conf/emnlp/ZhuLL0ZZ18} construct a large-scale corpus for a novel multimodal summarization task, which takes the news with images as input, and outputs a pictorial summary. They also present a multimodal attention method serving as a baseline for this task.

To further improve the quality of the model-selected images,~\citet{DBLP:conf/aaai/ZhuZZLZL20} propose to incorporate the multimodal reference into the training process, thus taking account of both summary generation and image selection to guide the abstractive model. LAMS~\cite{DBLP:conf/aaai/ZhangWSY21} considers image location as a critical factor for news summarization, and proposes a location-aware extractive approach for multimodal summarization.

Existing methods are only targeting either extractive or abstractive multimodal summarization. The tradeoff in the use of extractive and abstractive approaches has been discussed a lot in existing research \cite{DBLP:journals/corr/abs-1711-04434, DBLP:conf/aaai/JiaCSFYW21}: the former often achieve better factuality and efficiency, while the latter generates flexible and less redundant summaries. They both have merits in summarizing multimedia data.

We aim to propose a unified framework producing both extractive and abstractive summaries with a selected image preview that serves a multi-functional use. Our framework extends BART model \cite{DBLP:conf/acl/LewisLGGMLSZ20} to a unified multitask architecture.
The model comparison with related work is presented in Table~\ref{tab:Model_comp}, and more details are discussed in Section 4.3.
In addition, we distill the knowledge from state-of-the-art vision-language model CLIP \cite{DBLP:conf/icml/RadfordKHRGASAM21} to guide our training process, which avoids any requirement on the existence and quality of image captions.

\section{Methodology}
\subsection{Model Overview}

We propose a new multimodal summarization framework that unifies multiple subtasks of multimodal summarization. We use the BART model \cite{DBLP:conf/acl/LewisLGGMLSZ20} as our backbone architecture which consists of an encoder and decoder component. The overall architecture of our framework is shown in Fig.~\ref{fig:model}.

Given a multimodal document $D=\{T, V\}$, where $T=\{t_1,t_2,...,t_M\}$ is a sequence of $M$ tokens and $V=\{v_1,v_2,...,v_N\}$ is a collection of $N$ images, the proposed framework summarizes $D$ into a multimodal summary $Y=\{Y_t,Y_v\}$, where $Y_t=\{y_1,y_2,...,y_l\}$ denotes the textual summary limited by length $l$ and $Y_v$ is a subset of images extracted from the image collection $V$.

\subsection{Multimodal Reference Enhanced Encoder}

\subsubsection{Multimodal Encoder.}

Inspired by~\citet{DBLP:conf/icml/KimSK21}, we adopt a simple scheme, linear projection operating on image patches for obtaining visual embeddings. Specifically, each input image $v_i$ is sliced into patches and flattened as $\{v_{ik}\}$. Followed by linear projection and visual position embedding $v_{pos}$, $v_i$ is embedded into $e_{v_i}$. It is worth noting that, unlike existing methods~\cite{DBLP:conf/emnlp/ZhuLL0ZZ18,DBLP:conf/aaai/ZhuZZLZL20}, we take the first attempt to only use patch projection rather than any visual backbone.

Then, we extend the text encoder of BART to multimodal encoder, which takes the concatenation of text embeddings $e_t$ and visual embeddings $e_v$ as input and outputs their contextualized joint representations:
\begin{equation}
\begin{aligned}
& e_t = [t_{[\mathrm{CLS}]}; t_1 ; ...; t_{M}; t_{[\mathrm{SEP}]}]W_t , \\
& e_{v_i} = [v_{[\mathrm{CLS}]}; v_{i1} ; ...; v_{ik}]W_v + v_{pos} , \\
& e = [e_t; e_v] = [e_t; e_{v_i}; ...; e_{v_N}] + e_{pos} , \\
& h = [h_t; h_v] = f_{\mathrm{enc}}(e) , \\
\end{aligned}
\end{equation}
where $t_{[\mathrm{CLS}]}$ and $t_{[\mathrm{SEP}]}$ are the special token embeddings introduced to mark the start and the end of the text input, $e_{pos}$ is the multimodal sequence position embedding, $f_{\mathrm{enc}}$ denotes the encoder function, $W_t$ and $W_v$ are learnable parameters. Following~\citet{DBLP:conf/iclr/DosovitskiyB0WZ21}, we add $v_{[\mathrm{CLS}]}$ as the beginning token of each image, whose hidden state at the output of the encoder serves as the whole image representation.

\subsubsection{Knowledge Distillation for Image Reference.}

The multimodal summarization training set has only plain summary text reference, which lacks image reference for helping image selection in the training process.
Thus, the existing research~\cite{DBLP:conf/aaai/ZhuZZLZL20} considers \textbf{ROUGE-ranking} (RR) strategy to rank images and choose the top-k as image references, which thus extends text-only to multimodal reference.
In detail, RR uses the similarity of image caption and text reference to rank images.
However, this strategy relies on not only the presence of image captions but also the high relevance between images and their captions, which can lead to a very limited application scenario in reality.

Therefore, we adopt the Knowledge Distillation (KD) technique \cite{DBLP:journals/corr/HintonVD15} for distilling the text-image content relevance knowledge, which can get image reference without any image caption. Our proposed method aims to distill the knowledge from a teacher network, i.e., CLIP~\cite{DBLP:conf/icml/RadfordKHRGASAM21}, to a student network, i.e., our encoder. The student network is trained to mimic the behaviors of the teacher network, on calculating text-image content relevance score as well as ranking images according to the precedent score.

Specifically, we use the hidden state $h_{v_{[\mathrm{CLS}]}}$ of visual token \texttt{[CLS]} as the image representation $\overrightarrow{h_v}$, and feed it into a fully connected layer to predict a score:
\begin{equation}
g_v(\overrightarrow{h_v}) = W_v \overrightarrow{h_v} + b_v ,
\end{equation}
where $W_v$ and $b_v$ are learnable parameters.

Then, we employ CLIP~\cite{DBLP:conf/icml/RadfordKHRGASAM21} as a teacher model to calculate the cosine similarity scores between text embeddings generated by feeding textual summary $Y_t$ into its text encoder $\mathcal{T}$, and image embeddings generated by feeding image collection $V$ into its image encoder $\mathcal{V}$:
\begin{equation}
f(v, Y_t) = \mathrm{sim}(\mathcal{V}(v), \mathcal{T}(Y_t)) .
\end{equation}

To quantify the variation of ranking score distribution for our encoder and the CLIP model, we use the Kullback-Leibler (KL) divergence \cite{kullback1951information}.
Via distilling knowledge, our encoder intends to directly mimic the teacher model's score distribution.
Formally, CLIP-based KD can be modeled as minimizing the following objective functions with temperature $\tau$. Unless otherwise specified, all results reported in this paper use temperature $\tau=10$ which we find to perform best.
\begin{equation}
p_v(\overrightarrow{h_v}, \tau) = \frac{\mathrm{exp}(g_v(\overrightarrow{h_v}) / \tau)}{\sum_{v \in V} \mathrm{exp}(g_v(\overrightarrow{h_v}) / \tau)} ,
\end{equation}
\begin{equation}
q_v(v, Y_t, \tau) = \frac{\mathrm{exp}(f(v, Y_t) / \tau)}{\sum_{v \in V} \mathrm{exp}(f(v, Y_t) / \tau)} ,
\end{equation}
\begin{equation}
\mathcal{L}^{\mathrm{KD}} = \mathcal{L}_\mathrm{KL}(\mathbf{p} || \mathbf{q}) = - \sum_{v \in V} p_v \  \mathrm{ln} \frac{q_v}{p_v}.
\end{equation}

\subsubsection{Extractive Text Reference}
Due to the fact that if the encoder only obtains the reference from one single modality, the system may suffer from the modality-bias problem~\cite{DBLP:conf/aaai/ZhuZZLZL20}. Therefore, we introduce the extractive text reference to supervise our encoder. In other words, we regard extractive text summarization as one of the subtasks in our framework. We use a greedy algorithm similar to~\citet{DBLP:conf/aaai/NallapatiZZ17} to obtain an oracle summary within each document as the extractive text reference. In detail, our algorithm generates an oracle consisting of multiple sentences, which are selected greedily to maximize the ROUGE-L score against the gold summary.

In our encoder, for representing individual sentences, we insert extra \texttt{[CLS]} and \texttt{[SEP]} tokens at the start and end of each sentence, expecting each \texttt{[CLS]} token to collect features for the sentence following it, i.e., $\overrightarrow{h_t} = h_{t_{[\mathrm{CLS}]}}$. Thus, on top of which, a fully connected layer can be used to predict an extractive score for each sentence:
\begin{equation}
g_t(\overrightarrow{h_t}) = W_t \overrightarrow{h_t} + b_t ,
\end{equation}
where $W_t$ and $b_t$ are learnable parameters.

Given the extractive score per sentence, the regular binary cross-entropy loss is employed as the objective function for extractive text reference:
\begin{equation}
p_t(\overrightarrow{h_t}) = \frac{\mathrm{exp}(g_t(\overrightarrow{h_t}))}{\sum_t \mathrm{exp}(g_t(\overrightarrow{h_t}))} ,
\end{equation}
\begin{equation}
\mathcal{L}^{\mathrm{Ext}} = - \sum_t \mathrm{log}\ p_t(\overrightarrow{h_t}) .
\end{equation}

\subsection{Visual Guided Decoder}
Different from the original BART, our decoder has to attend to both the textual and visual content of the multimodal document instead of single modality input.
Inspired by the guidance-aware mechanism proposed by \citet{DBLP:conf/naacl/DouLHJN21}, which can introduce multiple textual guidance signals into the transformer decoder, we design a visual guided text generation decoder for better utilizing our encoded visual signals.
As shown in Fig.~\ref{fig:model}, after self-attention, the sequence in the decoder first attends to the encoded visual hidden states and generates the visual-guided cross-modal representation.
After that, this sequence continuously attends to the textual hidden states to produce the final representation:
\begin{equation}
\begin{aligned}
& y = \mathrm{LN}(y + \textrm{\large{C}}\textrm{\small{ROSS}}\textrm{\Large{A}}\textrm{\small{TTN}}(y; h_v)), \\
& y = \mathrm{LN}(y + \textrm{\large{C}}\textrm{\small{ROSS}}\textrm{\Large{A}}\textrm{\small{TTN}}(y; h_t)), \\
\end{aligned}
\end{equation}

Ideally, modeling the multimodal signals separately with two cross-attention blocks allows the decoder to be explicitly guided by signals from different modalities.
Overall, the decoder iteratively attends to previously generated tokens $y_{<j}$ and the encoder outputs $h$, and subsequently predicts the probability of future text tokens $p_y(y_j|y_{<j},h)$.
For multimodal conditional text generation task, i.e., multimodal abstractive summarization subtask, we train our model by minimizing the negative log-likelihood:
\begin{equation}
\mathcal{L}^{\mathrm{Abs}} = - \sum_{j=1}^{|y|} \mathrm{log}\ p_y (y_j|y_{<j},h) .
\end{equation}

Finally, the training loss $\mathcal{L}$ of our framework is a sum of the objectives of all these subtasks: image selection, extractive, and abstractive text summarization.
\begin{equation}
\mathcal{L} = \mathcal{L}^{\mathrm{KD}} + \mathcal{L}^{\mathrm{Ext}} + \mathcal{L}^{\mathrm{Abs}} .
\end{equation}

\section{Experiments}

\subsection{Datasets}

We use the MSMO dataset, which is collected by~\cite{DBLP:conf/emnlp/ZhuLL0ZZ18} for multimodal summarization.
It contains online news articles from the $Daily Mail$ website\footnote{http://www.dailymail.co.uk} paired with multiple images (6.58 images on average) and multi-sentence ground-truth summaries.
The dataset includes 293,965 training pairs, 10,355 validation pairs, and 10,261 test pairs.
In the test set, at most three images are annotated as a multimodal reference.

\subsection{Implementation Details}
Our framwork is built on 'bart-base'\footnote{https://huggingface.co/facebook/bart-base} version of BART~\cite{DBLP:conf/acl/LewisLGGMLSZ20} with its initialized parameters and tokenizer.
We use the released 'ViT-B-32'\footnote{https://github.com/openai/CLIP} version of CLIP~\cite{DBLP:conf/icml/RadfordKHRGASAM21} as the teacher network for knowledge distillation.
Meanwhile, following the image pre-processing used in CLIP, we resize the resolution of each image into 224$\times$224 with a patch projection yielding 49 patches.
We limit the number of images to 10 and pad the visual token \texttt{[CLS]} and image patch tokens together in a total length of 500 for batch training.
Besides, we truncate each article to 512 tokens, and the oracle summary in this paper is calculated on the truncated articles.
All models are trained for 30,000 steps with 750 steps for warm-up.
Model checkpoints are saved and evaluated on the validation set every 2,000 steps.
We select the top-3 checkpoints according to the validation loss and report the averaged results on the test set.

For abstractive summarization, we use beam search (size 5) in decoding, and tune the $\alpha$ for the length penalty~\cite{DBLP:journals/corr/WuSCLNMKCGMKSJL16} between 1.6 and 2.0 on the validation set; we decode until an end-of-sequence token is emitted.
For extractive summarization, we first use the model to obtain the score for each sentence. We then rank these sentences descendingly according to their scores and select the top-3 sentences as the summary.

We report the F1 \textbf{ROUGE} score via ROUGE-1.5.5.pl~\cite{lin2004rouge}, which calculates the overlap lexical units between generated and ground-truth sentences.
For image selection, we report image precision (\textbf{IP}) score, which is defined in~\cite{DBLP:conf/emnlp/ZhuLL0ZZ18} to represent whether an image is correctly selected as output.
Besides, \textbf{$\textrm{M}_{\textrm{sim}}$} is an image-text relevance metric which measures the maximum similarity between the output image and text summary via the cross-modal retrieval model, as stated in~\citet{DBLP:conf/emnlp/ZhuLL0ZZ18,DBLP:conf/aaai/ZhuZZLZL20}.

\begin{table}
\small
\centering
\begin{tabular}{p{3.4cm} | p{1.3cm}<{\centering} p{0.3cm}<{\centering} p{0.3cm}<{\centering} p{0.8cm}<{\centering}}
\toprule
Model                               & Multimodal & Ext.   & Abs.   & ImgSel    \\
\midrule
\midrule
BertExt~\citeyearpar{DBLP:conf/emnlp/LiuL19}            &           &\checkmark &           &      \\
BertAbs~\citeyearpar{DBLP:conf/emnlp/LiuL19}            &           &           &\checkmark &      \\
BertExtAbs~\citeyearpar{DBLP:conf/emnlp/LiuL19}         &           &\checkmark &\checkmark &      \\
BART~\citeyearpar{DBLP:conf/acl/LewisLGGMLSZ20}         &           &           &\checkmark & \\
\midrule
ATG/ATL/HAN~\citeyearpar{DBLP:conf/emnlp/ZhuLL0ZZ18}    &\checkmark &           &\checkmark &\checkmark \\
GR~\citeyearpar{DBLP:conf/emnlp/ZhuLL0ZZ18}             &\checkmark &\checkmark &           &\checkmark \\
MOF~\cite{DBLP:conf/aaai/ZhuZZLZL20}                    &\checkmark &           &\checkmark &\checkmark \\
LAMS~\cite{DBLP:conf/aaai/ZhangWSY21}                   &\checkmark &\checkmark &           & \\
\midrule
UniMS                                                   &\checkmark &\checkmark &\checkmark &\checkmark \\
\bottomrule
\end{tabular}
\caption{
Model comparison with baseline models.
Multimodal, Ext., Abs. and ImgSel denote whether the model could support multimodal input, perform extractive text summarization, abstractive text summarization, and image selection subtasks, respectively.
}
\label{tab:Model_comp}
\end{table}

\begin{table}
\small
\centering
\begin{tabular}{p{2.5cm} | p{0.7cm}<{\centering} p{0.7cm}<{\centering} p{0.7cm}<{\centering} p{0.7cm}<{\centering} p{0.7cm}<{\centering}}
\toprule
Model                       & R-1   & R-2   & R-L   & IP    & $\textrm{M}_{\textrm{sim}}$   \\
\midrule
\multicolumn{4}{c}{Text Abstractive} \\
\midrule
BertAbs~\citeyearpar{DBLP:conf/emnlp/LiuL19}    & 39.02 & 18.17 & 33.20 & - & - \\
BertExtAbs~\citeyearpar{DBLP:conf/emnlp/LiuL19} & 39.88 & 18.77 & 38.36 & - & - \\
BART~\citeyearpar{DBLP:conf/acl/LewisLGGMLSZ20} & \underline{41.83} & \underline{19.83} & \underline{39.74} & - & - \\
\midrule
\multicolumn{4}{c}{Multimodal Abstractive} \\
\midrule
ATG~\citeyearpar{DBLP:conf/emnlp/ZhuLL0ZZ18}*    & 40.63 & 18.12 & 37.53 & 59.28 & 25.82 \\
ATL~\citeyearpar{DBLP:conf/emnlp/ZhuLL0ZZ18}*    & 40.86 & 18.27 & 37.75 & 62.44 & 13.26 \\
HAN~\citeyearpar{DBLP:conf/emnlp/ZhuLL0ZZ18}*    & 40.82 & 18.30 & 37.70 & 61.83 & 12.22 \\
$\mathrm{MOF^{RR}_{enc}}$~\citeyearpar{DBLP:conf/aaai/ZhuZZLZL20}* & 41.05 & 18.29 & 37.74 & 62.63 & 26.23 \\
$\mathrm{MOF^{RR}_{dec}}$~\citeyearpar{DBLP:conf/aaai/ZhuZZLZL20}* & \underline{41.20} & 18.33 & 37.80 & \underline{65.45} & \underline{26.38} \\
\midrule
UniMS    & \textbf{42.94}    & \textbf{20.50}    & \textbf{40.96}    & \textbf{69.38} & \textbf{29.72} \\
\quad w/o Visual Guide                  & 42.71 & 20.26 & 40.76 & 69.14 & 29.64 \\
\quad w/o $\mathcal{L}^{\mathrm{Ext}}$  & 42.63 & 20.21 & 40.61 & 69.25 & 29.61 \\
\quad w/o Both                          & 42.55 & 19.88 & 40.14 & 69.22 & 29.57\\
UniMS-VL    & 40.81 & 18.83 & 38.83 & 68.26 & 29.45 \\
\bottomrule
\end{tabular}
\caption{
Experimental results for multimodal summarization on MSMO test set, where ``w/o Visual Guide" denotes to remove the visual-guided cross attention from the decoder and keep only one cross attention on all hidden states from two modalities, ``w/o $\mathcal{L}^{\mathrm{Ext}}$" denotes to the model without applying $\mathcal{L}^{\mathrm{Ext}}$ to the encoder and ``w/o Both" denotes to remove the former two.
Results marked by * are taken from the authors’ respective papers.
}
\label{tab:main}
\end{table}

\subsection{Baselines}
To show the effectiveness of our unified framework, we compare our model with existing text and multimodal summarization methods:

\noindent \textbf{BertSum}~\cite{DBLP:conf/emnlp/LiuL19} is an unified model for extractive and abstractive text summarization whose parameters are initialized with BERT~\cite{DBLP:conf/naacl/DevlinCLT19}. Its variants include \textbf{BertExt}, \textbf{BertAbs} and \textbf{BertExtAbs}.

\noindent \textbf{BART}~\cite{DBLP:conf/acl/LewisLGGMLSZ20} is the state-of-the-art abstractive pure text summarization model pretrained with a denoising autoencoding objective.

\noindent \textbf{ATG/ATL/HAN/GR} are proposed by~\citet{DBLP:conf/emnlp/ZhuLL0ZZ18}, the former three of which leverage the pointer generator network~\cite{DBLP:conf/acl/SeeLM17} for multimodal summarization via adopting the visual attention on global features, local features and hierarchically local features of images.
\textbf{GR} is an extractive method that ranks sentences and captions via LexRank~\cite{DBLP:journals/jair/ErkanR04} and guidance strategies.

\noindent \textbf{MOF variants}~\cite{DBLP:conf/aaai/ZhuZZLZL20} introduce the multimodal objective function with image reference obtained by ROUGE-ranking (\textbf{RR}) as described in Section 3.2.
Besides, they incorporate the last hidden states of the text encoder or the summary decoder into their proposed image discriminator and denote it as $\mathrm{MOF}_{\mathrm{enc}}$ and $\mathrm{MOF}_{\mathrm{dec}}$.

\noindent \textbf{LAMS}~\cite{DBLP:conf/aaai/ZhangWSY21} proposes a location-aware extractive approach to utilize the image locations for multimodal summarization.

\noindent \textbf{UniMS-VL}: ~\citet{DBLP:conf/icml/ChoLTB21} proposes VL-BART which extends BART text encoders to multimodal encoders by incoporating image region embeddings extracted by Fast-R-CNN~\cite{DBLP:journals/pami/RenHG017} as additional input.
VL-BART has shown good performances on vision-language tasks.
Therefore, we also aim to train our model with their released checkpoint as a starting point.

The comparison among our \textbf{UniMS} and different baseline models is shown in Table \ref{tab:Model_comp}, which demonstrates that our unified framework is able to cover all the subtasks of multimodal summarization.

\begin{table}
\small
\centering
\begin{tabular}{p{3.5cm} | p{0.7cm}<{\centering} p{0.7cm}<{\centering} p{0.7cm}<{\centering} p{0.7cm}<{\centering}}
\toprule
Model                       & R-1   & R-2   & R-L   \\
\midrule
ORACLE                      & 50.15 & 28.56 & 47.91 \\
LEAD-3                      & 39.94 & 18.56 & 38.38 \\
\midrule
GR~\citeyearpar{DBLP:conf/emnlp/ZhuLL0ZZ18}*     & 37.13 & 15.03 & 30.21 \\
BertExt~\citeyearpar{DBLP:conf/emnlp/LiuL19}    & 39.02 & 18.17 & 33.20 \\

LAMS-ATL~\citeyearpar{DBLP:conf/aaai/ZhangWSY21}* & 42.48 & 19.75 & 38.78 \\
LAMS-MFB~\citeyearpar{DBLP:conf/aaai/ZhangWSY21}* & \underline{\textbf{43.07}} & \underline{20.28} & \underline{39.34}    \\
\midrule
UniMS   & 42.58 & \textbf{20.29} & \textbf{40.91} \\
UniMS-VL    & 41.29 & 19.01 & 39.47 \\
\bottomrule

\end{tabular}
\caption{
Experimental results for extractive summarization on MSMO test set.
Results marked by * are taken from the respective papers.
}
\label{tab:extractive}
\end{table}

\begin{table}
\small
\centering
\begin{tabular}{p{2.2cm} | p{0.7cm}<{\centering} p{0.7cm}<{\centering} p{0.7cm}<{\centering}}
\toprule
Model                       & R-L     & IP    & $\textrm{M}_{\textrm{sim}}$   \\ %
\midrule
ROUGE-ranking   & 40.90 & 69.25 & 29.56 \\
UniMS           & \textbf{40.96} & \textbf{69.38} & \textbf{29.72} \\
\bottomrule
\end{tabular}
\caption{
Experimental results for multimodal summarization with different methods of building image references.
}
\label{tab:imageref}
\end{table}

\subsection{Automatic Evaluation}

Table~\ref{tab:main} summarizes the results on abstractive summarization and image selection subtasks. The first block in the table includes abstractive summarization methods with text-only input, while the second block includes abstractive methods with multimodal input.
We find that, by fine-tuning the pre-trained BART model \cite{DBLP:conf/acl/LewisLGGMLSZ20}, the results of abstractive text summarization is able to achieve competitive performances as state-of-the-art models that additionally use visual signals, which indicates the powerful language modeling capabilities of BART and motivates us to extend BART to handle multi-modalities.
Our proposed \textbf{UniMS} outperforms state-of-the-art methods in both abstractive summarization and image selection tasks, demonstrating the superiority of our multimodal summarization model.

In addition, our model can achieve extractive goals with $\mathcal{L}^{\mathrm{Ext}}$.
We display the results of our extractive summarization in Table~\ref{tab:extractive}. As an upper bound, the table begins with an extractive ORACLE result as described in Section 3.2. We also present the LEAD-3 baseline, which simply selects the first three sentences from a document as extractive output.
The second block includes various extractive summarization methods with text-only input, i.e., BertExt, and multimodal input, i.e., LAMS. Our extractive model can achieve comparable performance to the state-of-the-art extractive method \cite{DBLP:conf/aaai/ZhangWSY21} that additionally models human-labeled image locations as input.

Overall, our proposed unified multitask framework provides superior performances in all subtasks of multimodal summarization.

\begin{table}[t]
\small
\centering
\begin{tabular}{p{2.1cm} | p{0.6cm}<{\centering} p{0.6cm}<{\centering} p{0.6cm}<{\centering} p{0.9cm}<{\centering}}
\toprule
Model & R-L & IP & $\textrm{M}_{\textrm{sim}}$ & Para.$\uparrow$\% \\ %
\midrule

LinearProjection    & \textbf{40.96}   & \textbf{69.38} & \textbf{29.72}  & \textbf{10.07}  \\ %
\midrule
ResNet50      & 40.79 & 68.92 & \underline{29.66} & 30.22 \\ %
CLIP-ResNet50 & 40.83 & 69.14 & 23.73 & \underline{28.78} \\ %
CLIP-ViT-B-32 & \underline{40.86} & \underline{69.16} & 29.62 & 74.10 \\ %
\midrule
Fast-RCNN     & 40.76 & 68.33 & 29.62 & 41.73 \\ %
UniMS-VL      & 38.83 & 68.26 & 29.45 & 41.73 \\ %
\bottomrule
\end{tabular}
\caption{
Experimental results for multimodal summarization with different visual backbones.
}
\label{tab:visual}
\vspace{-0.3cm}
\end{table}

\begin{figure}[t]
\centering
\includegraphics[width=.5\textwidth]{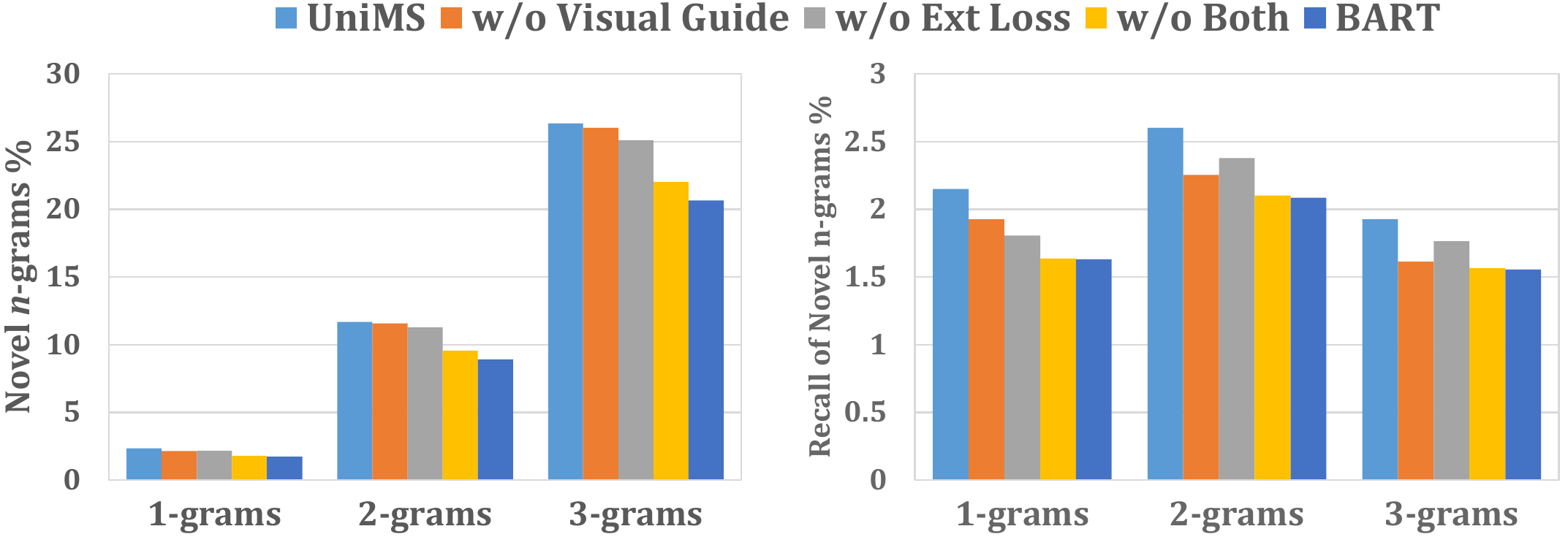}
\caption{
Novel $n$-grams and the recall of novel $n$-grams appearing in the ground-truth reference.
Introducing the visual guided decoder and extractive objective can enhance the ability of our framework to generate more novel $n$-grams.
}
\label{fig:novelngram}
\end{figure}

\begin{figure*}
\centering
\includegraphics[width=\textwidth]{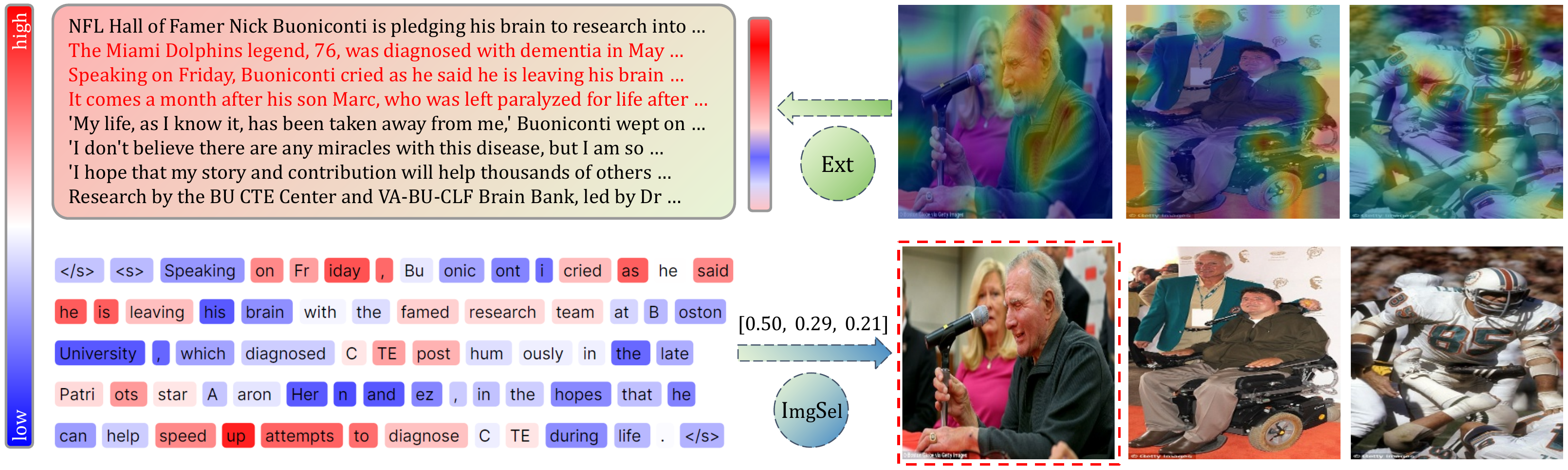}
\caption{
Visualizations for indicating how one modality affects the other in our multimodal encoder. Note that the importance scores decrease from red to blue. The red three sentences are extractive results and the first image with a red border owns the highest score 0.5 in the image selection. Our encoder is able to recognize summary reference related tokens as important, such as the person regions from images, and verb tokens like "\texttt{said}" and "\texttt{attempt}".
}
\label{fig:case}
\end{figure*}

\subsection{Model Analysis}
\subsubsection{Ablation study.}
Table \ref{tab:main} also shows the ablation test of our model when some components are removed.
It shows the degradation in performances of abstractive summarization and image selection tasks when removing the corresponding components.
Meanwhile, we find that training from the BART pretraining parameters is better than the VL-BART.
It may be because the pre-training task of VL-BART is primarily designed for vision-language tasks, such as VQA, etc.
Therefore, the original BART's capability, such as summarization, is corrupted during their training process.
We believe using a visual-languaged pretrained model particularly designed for multimodal summarization tasks could further improve our framework, which we leave as future work.

\subsubsection{Image Reference.}
Table~\ref{tab:imageref} depicts the model performances varying with different strategies of building image references. We adopt the ROUGE-ranking (\textbf{RR}) strategy from MOF~\cite{DBLP:conf/aaai/ZhuZZLZL20} to our framework.
Results show that RR is a strong competitor since the high-quality image caption of MSMO dataset has already contained condensed textual information of visual modalities. However, RR relies on the presence of high-quality image captions, while our KD method is more flexible with no additional dependency on captions, leading to wide use in reality.

\begin{table}[t]
\small
\centering
\begin{tabular}{p{2.5cm} | p{0.9cm}<{\centering} p{0.9cm}<{\centering} p{0.9cm}<{\centering} p{0.9cm}<{\centering}}
\toprule
Models  & \multicolumn{2}{c}{Abs.} & Ext. & ImgSel \\
        & Consist. & Relev. & Relev. & Relev. \\
\midrule
BART                                    & 2.18 & 2.40 & -     & - \\
UniMS                                   & \textbf{2.32} & \textbf{2.45} & 2.37  & \underline{\textbf{2.44}} \\
\quad w/o Visual Guide                  & 2.26 & 2.42 & \textbf{2.41}  & \underline{\textbf{2.44}} \\
\quad w/o $\mathcal{L}^{\mathrm{Ext}}$  & 2.22 & 2.40 & -     & 2.42 \\
\quad w/o Both                          & 2.19 & 2.39 & -     & 2.38 \\
\bottomrule
\end{tabular}
\caption{
Human evaluation of different model outputs.
All the improvements are significant ($p<0.001$).
}
\label{tab:human}
\end{table}

\subsubsection{Visual Backbone.}
To study the impact of visual backbones in our framework, we conduct an experiment on evaluating the model performances vary with employing different visual backbones, as shown in Table~\ref{tab:visual}.
In terms of grid features, we use pretrained visual backbones, i.e., ResNet50, CLIP-ResNet50, and CLIP-ViT-B-32, to extract the 49 image patch features.
In terms of region features, we represent an input image $v$ with $n=49$ object regions with Faster R-CNN~\cite{DBLP:journals/pami/RenHG017} pretrained on Visual Genome~\cite{DBLP:conf/cvpr/00010BT0GZ18}.

Results show that the pretrained visual backbones cannot significantly improve the performance of our model. Meanwhile, the use of pretrained visual backbones increases the overhead of obtaining visual embeddings. Using the linear projection only brings an additional $10\%$ gain in parameters compared to BART. As a result, we choose linear projection to obtain visual embeddings in our framework.

\subsubsection{Novel $n$-grams.}
Following~\citet{DBLP:conf/naacl/DouLHJN21}, we count the number of novel $n$-grams in the output summaries, namely the newly generated $n$-grams that do not appear in the source document.
As shown in Fig.~\ref{fig:novelngram}, all the variants of our model generate more novel $n$-grams and cover more novel $n$-grams that exist in the ground-truth references than the baseline BART. It indicates that our modifications on top of BART can indeed generate more novel expressions in the abstractive summary.
Additionally, it is clear from Fig.~\ref{fig:novelngram} that introducing the extractive loss and visual guided decoder brings a significant improvement on enabling models to generate more novel $n$-grams.

\subsubsection{Human Evaluation.}

We further conduct a manual evaluation to assess the quality of the generated multimodal summarization. We randomly sample 200 data points from the test set and recruit three people from Amazon Mechanical Turk to rate them between 1 to 3 points on multiple qualitative aspects.
For extractive summarization, we ask the annotators to evaluate the relevance (i.e., Relev.), which measures whether the summary captures the key points of the source document.
For abstractive summarization, we ask the annotators to evaluate the consistency (i.e., Consist.), which measures the factual alignment between the summary and the source document.
In particular for image selection, we ask the annotators to measure whether the summary captures the key points of the selected image as the relevance (i.e., Relev.) score, which indicates the image-text relevance of multimodal output. Table~\ref{tab:human} shows that our framework can generate more faithful and relevant summaries compared to other variants and baselines.

\subsubsection{Visualizations}

It is of interest to visualize how one modality affects the other in our multimodal encoder.
As shown in Fig.~\ref{fig:case}, we adopt the class activation mapping (CAM)~\cite{DBLP:conf/cvpr/ZhouKLOT16} technique to visualize the importance scores by feeding the last hidden states of our encoder to the fully connected layers used separately in extractive summarization and image selection tasks.
Specifically, for extractive summarization, we employ its fully connected layer on the last hidden states of visual embeddings to indicate which region of images is regarded as important by this layer.
Similarly for image selection, we employ its fully connected layer on the last hidden states of textual tokens to indicate which set of tokens is regarded as important by this image selection layer.
As shown in Fig. \ref{fig:case}, we find that our framework surprisingly attends to the person and player region from images, and the verb tokens such as ``\texttt{said}" and ``\texttt{attempt}", which is quite close to the key points of summary reference.

\section{Conclusion}
We propose a unified framework for multimodal summarization that is able to produce extractive and abstractive text output, and jointly select image output for the final pictorial summary. Grounding on BART, our method uses an encoder-decoder architecture consisting of three training objectives, i.e., extractive \& abstractive text summarization, and image selection. In particular, extractive and image selection objectives cooperatively supervise the encoder. To improve the image selection, we distill the knowledge from existing pre-trained vision-language models. We also propose a visual guided decoder that separately attends to the textual and visual modalities while performing abstractive text generation. Overall, our best model achieves a new state-of-the-art result on the MSMO dataset.
We believe such a unified
framework
with multi-functional use can serve as a stepping stone to further improve multimodal summarization, as well as baselines against which future methods are tested.

\bibliography{aaai22, mybib}

\begin{thebibliography}{27}
\providecommand{\natexlab}[1]{#1}

\bibitem[{Anderson et~al.(2018)Anderson, He, Buehler, Teney, Johnson, Gould,
  and Zhang}]{DBLP:conf/cvpr/00010BT0GZ18}
Anderson, P.; He, X.; Buehler, C.; Teney, D.; Johnson, M.; Gould, S.; and
  Zhang, L. 2018.
\newblock Bottom-Up and Top-Down Attention for Image Captioning and Visual
  Question Answering.
\newblock In \emph{CVPR}, 6077--6086.

\bibitem[{Baltrusaitis, Ahuja, and
  Morency(2019)}]{DBLP:journals/pami/BaltrusaitisAM19}
Baltrusaitis, T.; Ahuja, C.; and Morency, L. 2019.
\newblock Multimodal Machine Learning: {A} Survey and Taxonomy.
\newblock \emph{{IEEE} Trans. Pattern Anal. Mach. Intell.}, 41(2): 423--443.

\bibitem[{Cao et~al.(2017)Cao, Wei, Li, and
  Li}]{DBLP:journals/corr/abs-1711-04434}
Cao, Z.; Wei, F.; Li, W.; and Li, S. 2017.
\newblock Faithful to the Original: Fact Aware Neural Abstractive
  Summarization.
\newblock \emph{CoRR}, abs/1711.04434.

\bibitem[{Cho et~al.(2021)Cho, Lei, Tan, and Bansal}]{DBLP:conf/icml/ChoLTB21}
Cho, J.; Lei, J.; Tan, H.; and Bansal, M. 2021.
\newblock Unifying Vision-and-Language Tasks via Text Generation.
\newblock In \emph{ICML}, volume 139, 1931--1942.

\bibitem[{Devlin et~al.(2019)Devlin, Chang, Lee, and
  Toutanova}]{DBLP:conf/naacl/DevlinCLT19}
Devlin, J.; Chang, M.; Lee, K.; and Toutanova, K. 2019.
\newblock {BERT:} Pre-training of Deep Bidirectional Transformers for Language
  Understanding.
\newblock In \emph{NAACL-HLT}, 4171--4186.

\bibitem[{Dosovitskiy et~al.(2021)Dosovitskiy, Beyer, Kolesnikov, Weissenborn,
  Zhai, Unterthiner, Dehghani, Minderer, Heigold, Gelly, Uszkoreit, and
  Houlsby}]{DBLP:conf/iclr/DosovitskiyB0WZ21}
Dosovitskiy, A.; Beyer, L.; Kolesnikov, A.; Weissenborn, D.; Zhai, X.;
  Unterthiner, T.; Dehghani, M.; Minderer, M.; Heigold, G.; Gelly, S.;
  Uszkoreit, J.; and Houlsby, N. 2021.
\newblock An Image is Worth 16x16 Words: Transformers for Image Recognition at
  Scale.
\newblock In \emph{ICLR}.

\bibitem[{Dou et~al.(2021)Dou, Liu, Hayashi, Jiang, and
  Neubig}]{DBLP:conf/naacl/DouLHJN21}
Dou, Z.; Liu, P.; Hayashi, H.; Jiang, Z.; and Neubig, G. 2021.
\newblock GSum: {A} General Framework for Guided Neural Abstractive
  Summarization.
\newblock In \emph{NAACL-HLT}, 4830--4842.

\bibitem[{Erkan and Radev(2004)}]{DBLP:journals/jair/ErkanR04}
Erkan, G.; and Radev, D.~R. 2004.
\newblock LexRank: Graph-based Lexical Centrality as Salience in Text
  Summarization.
\newblock \emph{J. Artif. Intell. Res.}, 22: 457--479.

\bibitem[{He and Deng(2017)}]{DBLP:journals/spm/HeD17}
He, X.; and Deng, L. 2017.
\newblock Deep Learning for Image-to-Text Generation: {A} Technical Overview.
\newblock \emph{{IEEE} Signal Process. Mag.}, 34(6): 109--116.

\bibitem[{Hinton, Vinyals, and Dean(2015)}]{DBLP:journals/corr/HintonVD15}
Hinton, G.~E.; Vinyals, O.; and Dean, J. 2015.
\newblock Distilling the Knowledge in a Neural Network.
\newblock \emph{CoRR}, abs/1503.02531.

\bibitem[{Jia et~al.(2021)Jia, Cao, Shi, Fang, Yin, and
  Wang}]{DBLP:conf/aaai/JiaCSFYW21}
Jia, R.; Cao, Y.; Shi, H.; Fang, F.; Yin, P.; and Wang, S. 2021.
\newblock Flexible Non-Autoregressive Extractive Summarization with Threshold:
  How to Extract a Non-Fixed Number of Summary Sentences.
\newblock In \emph{AAAI}, 13134--13142.

\bibitem[{Kim, Son, and Kim(2021)}]{DBLP:conf/icml/KimSK21}
Kim, W.; Son, B.; and Kim, I. 2021.
\newblock ViLT: Vision-and-Language Transformer Without Convolution or Region
  Supervision.
\newblock In \emph{ICML}, volume 139, 5583--5594.

\bibitem[{Kullback and Leibler(1951)}]{kullback1951information}
Kullback, S.; and Leibler, R.~A. 1951.
\newblock On information and sufficiency.
\newblock \emph{The annals of mathematical statistics}, 22(1): 79--86.

\bibitem[{Lewis et~al.(2020)Lewis, Liu, Goyal, Ghazvininejad, Mohamed, Levy,
  Stoyanov, and Zettlemoyer}]{DBLP:conf/acl/LewisLGGMLSZ20}
Lewis, M.; Liu, Y.; Goyal, N.; Ghazvininejad, M.; Mohamed, A.; Levy, O.;
  Stoyanov, V.; and Zettlemoyer, L. 2020.
\newblock {BART:} Denoising Sequence-to-Sequence Pre-training for Natural
  Language Generation, Translation, and Comprehension.
\newblock In \emph{ACL}, 7871--7880.

\bibitem[{Li et~al.(2017)Li, Zhu, Ma, Zhang, and
  Zong}]{DBLP:conf/emnlp/LiZMZZ17}
Li, H.; Zhu, J.; Ma, C.; Zhang, J.; and Zong, C. 2017.
\newblock Multi-modal Summarization for Asynchronous Collection of Text, Image,
  Audio and Video.
\newblock In \emph{EMNLP}, 1092--1102.

\bibitem[{Lin(2004)}]{lin2004rouge}
Lin, C.-Y. 2004.
\newblock Rouge: A package for automatic evaluation of summaries.
\newblock In \emph{Text summarization branches out}, 74--81.

\bibitem[{Liu and Lapata(2019)}]{DBLP:conf/emnlp/LiuL19}
Liu, Y.; and Lapata, M. 2019.
\newblock Text Summarization with Pretrained Encoders.
\newblock In \emph{EMNLP-IJCNLP}, 3728--3738.

\bibitem[{Nallapati, Zhai, and Zhou(2017)}]{DBLP:conf/aaai/NallapatiZZ17}
Nallapati, R.; Zhai, F.; and Zhou, B. 2017.
\newblock SummaRuNNer: {A} Recurrent Neural Network Based Sequence Model for
  Extractive Summarization of Documents.
\newblock In \emph{AAAI}, 3075--3081.

\bibitem[{Radford et~al.(2021)Radford, Kim, Hallacy, Ramesh, Goh, Agarwal,
  Sastry, Askell, Mishkin, Clark, Krueger, and
  Sutskever}]{DBLP:conf/icml/RadfordKHRGASAM21}
Radford, A.; Kim, J.~W.; Hallacy, C.; Ramesh, A.; Goh, G.; Agarwal, S.; Sastry,
  G.; Askell, A.; Mishkin, P.; Clark, J.; Krueger, G.; and Sutskever, I. 2021.
\newblock Learning Transferable Visual Models From Natural Language
  Supervision.
\newblock In \emph{ICML}, volume 139, 8748--8763.

\bibitem[{Ren et~al.(2017)Ren, He, Girshick, and
  Sun}]{DBLP:journals/pami/RenHG017}
Ren, S.; He, K.; Girshick, R.~B.; and Sun, J. 2017.
\newblock Faster {R-CNN:} Towards Real-Time Object Detection with Region
  Proposal Networks.
\newblock \emph{{IEEE} Trans. Pattern Anal. Mach. Intell.}, 39(6): 1137--1149.

\bibitem[{See, Liu, and Manning(2017)}]{DBLP:conf/acl/SeeLM17}
See, A.; Liu, P.~J.; and Manning, C.~D. 2017.
\newblock Get To The Point: Summarization with Pointer-Generator Networks.
\newblock In \emph{ACL}, 1073--1083.

\bibitem[{Srihari(1994)}]{DBLP:journals/air/Srihari94}
Srihari, R.~K. 1994.
\newblock Computational Models for Integrating Linguistic and Visual
  Information: {A} Survey.
\newblock \emph{Artif. Intell. Rev.}, 8(5-6): 349--369.

\bibitem[{Wu et~al.(2016)Wu, Schuster, Chen, Le, Norouzi, Macherey, Krikun,
  Cao, Gao, Macherey, Klingner, Shah, Johnson, Liu, Kaiser, Gouws, Kato, Kudo,
  Kazawa, Stevens, Kurian, Patil, Wang, Young, Smith, Riesa, Rudnick, Vinyals,
  Corrado, Hughes, and Dean}]{DBLP:journals/corr/WuSCLNMKCGMKSJL16}
Wu, Y.; Schuster, M.; Chen, Z.; Le, Q.~V.; Norouzi, M.; Macherey, W.; Krikun,
  M.; Cao, Y.; Gao, Q.; Macherey, K.; Klingner, J.; Shah, A.; Johnson, M.; Liu,
  X.; Kaiser, L.; Gouws, S.; Kato, Y.; Kudo, T.; Kazawa, H.; Stevens, K.;
  Kurian, G.; Patil, N.; Wang, W.; Young, C.; Smith, J.; Riesa, J.; Rudnick,
  A.; Vinyals, O.; Corrado, G.; Hughes, M.; and Dean, J. 2016.
\newblock Google's Neural Machine Translation System: Bridging the Gap between
  Human and Machine Translation.
\newblock \emph{CoRR}, abs/1609.08144.

\bibitem[{Zhang et~al.(2021)Zhang, Wang, Sun, and
  Yang}]{DBLP:conf/aaai/ZhangWSY21}
Zhang, Z.; Wang, J.; Sun, Z.; and Yang, Z. 2021.
\newblock {LAMS:} {A} Location-aware Approach for Multimodal Summarization
  (Student Abstract).
\newblock In \emph{AAAI}, 15949--15950.

\bibitem[{Zhou et~al.(2016)Zhou, Khosla, Lapedriza, Oliva, and
  Torralba}]{DBLP:conf/cvpr/ZhouKLOT16}
Zhou, B.; Khosla, A.; Lapedriza, {\`{A}}.; Oliva, A.; and Torralba, A. 2016.
\newblock Learning Deep Features for Discriminative Localization.
\newblock In \emph{CVPR}, 2921--2929.

\bibitem[{Zhu et~al.(2018)Zhu, Li, Liu, Zhou, Zhang, and
  Zong}]{DBLP:conf/emnlp/ZhuLL0ZZ18}
Zhu, J.; Li, H.; Liu, T.; Zhou, Y.; Zhang, J.; and Zong, C. 2018.
\newblock {MSMO:} Multimodal Summarization with Multimodal Output.
\newblock In \emph{EMNLP}, 4154--4164.

\bibitem[{Zhu et~al.(2020)Zhu, Zhou, Zhang, Li, Zong, and
  Li}]{DBLP:conf/aaai/ZhuZZLZL20}
Zhu, J.; Zhou, Y.; Zhang, J.; Li, H.; Zong, C.; and Li, C. 2020.
\newblock Multimodal Summarization with Guidance of Multimodal Reference.
\newblock In \emph{AAAI}, 9749--9756.

\end{thebibliography}

\clearpage
\appendix

\section{Appendix: Further Discussion}

\subsubsection{Distillation Temperature.}

We also study the model performances varying with the temperature $\tau$ used in knowledge distillation, as shown in Table~\ref{tab:temperature}.
Our model achieves the best performance at $\tau = 10$.
Analysis of the experimental losses shows that when $\tau$ is small, the model quickly reaches a high IP (image precision) on image selection at the very beginning of training, and then overfits. In contrast, a high $\tau$ prevents overfitting and ultimately results in a better performance across all subtasks. As mentioned, we use $\tau=10$ in our manuscript.

\subsubsection{Encoder Reference Layer.}
We further explore where the image reference should be added to the encoder, i.e. on which layer of the transformer encoder.
Kindly note that the $\tau=1$ is used in this exploratory experiment, which is different from the main experiment ($\tau=10$) in our manuscript. We have not explored it across all variants of $\tau$ given the computational cost, which we would like leave as future work.
As shown in Table~\ref{tab:layer},
adding the image reference associated with its knowledge distillation technique to the front layers improves abstractive summarization. We believe adopting it to the front layers might inject visual signals to affect textual features and generation at an early stage. However, the performance on the image selection task reaches the best at middle layers ($\textrm{L}=3$ or $4$). It shows the model requires deeper layers to integrate textual and visual modalities to perform selection tasks. Note that in the manuscript, we finally use the $\textrm{L}=6$ as a more standard setting on top of BART. We believe our framework with less layers and parameters is still able to achieve competitive performances.

\begin{table}
\small
\centering
\begin{tabular}{p{2.2cm} | p{0.7cm}<{\centering} p{0.7cm}<{\centering} p{0.7cm}<{\centering}}
\toprule
Model                       & R-L     & IP    & $\textrm{M}_{\textrm{sim}}$   \\ %
\midrule
UniMS  w/ $\tau=1$  & 40.74 & 68.82 & 29.66 \\ %
\qquad w/ $\tau=5$  & 40.87 & 69.09 & 29.69 \\ %
\qquad w/ $\tau=10$ & \textbf{40.96} & \textbf{69.38} & \textbf{29.72} \\ %
\qquad w/ $\tau=20$ & 40.84 & 69.19 & 29.67 \\ %
\bottomrule
\end{tabular}
\caption{
Experimental results for multimodal summarization with different temperatures $\tau$.
}
\label{tab:temperature}
\end{table}

\begin{table}
\small
\centering
\begin{tabular}{p{2cm} | p{0.7cm}<{\centering} p{0.7cm}<{\centering} p{0.7cm}<{\centering} p{0.7cm}<{\centering}} %
\toprule
Model   & R-1   & R-2   & R-L   & IP  \\ %
\midrule
$\mathcal{L}^{\mathrm{KD}}$ w/ $\textrm{L}=1$  & 42.82 & 20.38 & 40.85 & 69.19 \\
\qquad w/ $\textrm{L}=2$ & \textbf{42.83} & 20.37 & \textbf{40.94} & 67.98 \\
\qquad w/ $\textrm{L}=3$ & 42.78 & \textbf{20.41} & 40.87 & 69.24 \\
\qquad w/ $\textrm{L}=4$ & 42.71 & 20.31 & 40.73 & \textbf{69.35} \\
\qquad w/ $\textrm{L}=5$ & 42.72 & 20.20 & 40.80 & 67.51 \\
\qquad w/ $\textrm{L}=6$ & 42.70 & 20.27 & 40.74 & 68.82 \\
\bottomrule
\end{tabular}
\caption{
Experimental results with different layers with distillation temperature $\tau=1$.
}
\label{tab:layer}
\end{table}

\end{document}